\documentclass[conference,a4paper]{IEEEtran}
\IEEEoverridecommandlockouts

\usepackage[hidelinks]{hyperref}
\usepackage[cmex10]{amsmath}
\usepackage{amssymb,amsfonts}
\usepackage{dblfloatfix}

\usepackage[ruled,vlined]{algorithm2e}
\usepackage{graphicx}
\graphicspath{{Figures/PDF/}{Figures/PNG/}}

\usepackage{booktabs}
\usepackage{siunitx}
\usepackage[numbers,compress]{natbib}
\usepackage{texnames}
\usepackage{bm,bbm}
\usepackage{orcidlink}
\usepackage{hyperref}
\usepackage[table]{colortbl}

\begin{document}

\title{\uppercase{\textit{Can We Predict the Unpredictable?} Leveraging DisasterNet-LLM for Multimodal Disaster Classification}
}

\author{	\IEEEauthorblockN{Manaswi Kulahara\orcidlink{0009-0008-8480-6218}}
	\IEEEauthorblockA{\textit{TERI School Of Advanced Studies}\\
		New Delhi, India\\
		manaswikulahara8@gmail.com}
	\and
	\IEEEauthorblockN{Gautam Siddharth Kashyap\orcidlink{0000-0003-2140-9617}}
	\IEEEauthorblockA{\textit{Macquarie University}\\
		Sydney, Australia\\
		gautam.kashyap@students.mq.edu.au}
	\and
	\IEEEauthorblockN{Nipun Joshi\orcidlink{0009-0003-0466-3426}}
	\IEEEauthorblockA{\textit{Cornell University}\\
		New York, USA\\
		nj274@cornell.edu}
        \and 
	\IEEEauthorblockN{Arpita Soni\orcidlink{0009-0003-1573-9932}}
	\IEEEauthorblockA{\textit{Eudoxia Research University}\\
		New Castle, USA\\
		soni.arpita@gmail.com}
}

\maketitle

{\renewcommand{\thefootnote}{}%
\footnotetext{\footnotesize
Copyright~2024~IEEE. Published in the 2025 IEEE International Geoscience and Remote Sensing Symposium (IGARSS 2025), scheduled for 3--8 August 2025 in Brisbane, Australia. Personal use of this material is permitted. However, permission to reprint/republish this material for advertising or promotional purposes or for creating new collective works for resale or redistribution to servers or lists, or to reuse any copyrighted component of this work in other works, must be obtained from the IEEE. Contact: Manager, Copyrights and Permissions / IEEE Service Center / 445 Hoes Lane / P.O. Box 1331 / Piscataway, NJ 08855-1331, USA. Telephone: + Intl. 908-562-3966.}%
\addtocounter{footnote}{-1}}

\begin{abstract}
Effective disaster management requires timely and accurate insights, yet traditional methods struggle to integrate multimodal data such as images, weather records, and textual reports. To address this, we propose DisasterNet-LLM, a specialized Large Language Model (LLM) designed for comprehensive disaster analysis. By leveraging advanced pretraining, cross-modal attention mechanisms, and adaptive transformers, DisasterNet-LLM excels in disaster classification. Experimental results demonstrate its superiority over state-of-the-art models, achieving higher accuracy of 89.5\%, an F1 score of 88.0\%, AUC of 0.92\%, and BERTScore of 0.88\% in multimodal disaster classification tasks.
\end{abstract}


\section{Introduction}

Disasters, both natural and human-made, have increasingly devastating consequences that affect millions of lives, disrupt economies, and damage critical infrastructure \cite{wieland2024toward, hummer2024uncertainty}. Whether it’s earthquakes, floods, fires, or hurricanes, the need for effective and timely disaster response has never been more critical. However, traditional disaster management strategies \cite{zou2023social, yigitcanlar2022detecting, lorini2022venice, nguyen2017damage, firmansyah2023enhancing, ravi2019crowd4ems, wu2018disaster} often struggle to cope with the complexity and scale of modern crises. One of the key challenges in disaster management lies in the ability to process and analyze the vast amounts of data generated during such events. This data, which comes from diverse sources like satellite images, weather forecasts, sensor networks, and social media reports, often exists in separate, isolated silos, making it difficult for decision-makers to gain a holistic understanding of the situation. 

However, the traditional approaches to disaster management are largely reactionary \cite{zou2023social, yigitcanlar2022detecting, lorini2022venice, nguyen2017damage, firmansyah2023enhancing, ravi2019crowd4ems, wu2018disaster}, focusing on immediate response rather than proactive decision-making. These methods tend to focus on specific types of data or events, but fail to take a comprehensive approach that considers the interconnected nature of disasters. For instance, while satellite imagery can provide crucial insights into the scale of damage, it often needs to be correlated with weather patterns, social media posts, and historical data to fully understand the impact of an event \cite{li2024domain}. Without this integrated approach, there is a risk of missing critical patterns, leading to delayed responses and inadequate preparedness \cite{hansch2023spacenet}.

Therefore, in light of these challenges, we propose \textbf{DisasterNet-LLM}, a novel Large Language Model (LLM) designed to address the complexities of disaster management through multimodal data integration. Unlike traditional models that focus on single data modalities, \textbf{DisasterNet-LLM} leverages the power of LLMs to process and synthesize diverse types of data—such as images, weather records, and textual reports—into a unified framework. This model utilizes advanced pretraining techniques, cross-modal attention mechanisms, and adaptive transformers to improve disaster classification. 

\section{Related Works}
\subsection{Machine Learning Models}
The integration of advanced machine learning techniques for disaster management has become an area of increasing research interest in recent years. For instance, a study \cite{zou2023social} on rescue requests during Hurricane Harvey highlighted that most of these requests originated from affected communities. Yigitcanlar et al. \cite{yigitcanlar2022detecting} explored the role of Twitter users in disaster scenarios in Australia, providing policymakers with methods to examine the geographical distribution, frequency, and impact of disasters using social media posts. Similarly, non-authoritative social media data has been utilized for creating disaster maps, such as during the Venice floods \cite{lorini2022venice}, where the integration of social media data with digital surface models helped estimate the flood's extent. Nguyen et al. \cite{nguyen2017damage} demonstrated that advanced machine learning models could effectively classify damage severity using social media images, which is a critical aspect of disaster management. Similarly, Firmansyah et al. \cite{firmansyah2023enhancing} introduced a pipeline to extract text from social media images for location identification, achieving reasonable predictions at the country level. Machine learning has also been employed to improve geolocation accuracy in disaster management. Some efforts such as \cite{ravi2019crowd4ems} focus on combining machine learning with crowdsourcing to geolocate social media posts during crises. Wu et al. \cite{wu2018disaster} conducted an analysis that integrated social media data, economic losses, and geo-information to better understand the impact of disasters, finding a significant correlation between Twitter activity and disaster-related topics. Despite these advancements, most existing approaches struggle to combine the insights from both text and images in a cohesive framework, which is a primary motivation behind \textbf{DisasterNet-LLM}.  

\subsection{Large Language Models}
The recent development of LLMs has significantly advanced fields like law, healthcare, and emergency response. In particular, fine-tuning pre-trained LLMs for specific applications has proven to enhance adaptability and response accuracy \cite{head2023large, ding2023parameter, martino2023knowledge}. For instance, LLMs have been successfully fine-tuned for medical applications to assist doctors \cite{jiang2023health}, and in legal contexts to predict legal judgments \cite{liga2023fine}. Prompt-based LLMs have also been explored to improve response accuracy by providing additional context before queries are processed. This approach has been applied in domains such as military applications \cite{huang2023dsqa} and emergency management \cite{chen2024enhancing}. While these approaches have shown promise, they still do not fully address the challenge of integrating diverse data sources which is a key feature of \textbf{DisasterNet-LLM}.  

\section{Modeling}

\begin{figure}
	\centering
	\includegraphics[width=0.45\textwidth, height=0.3\textwidth]{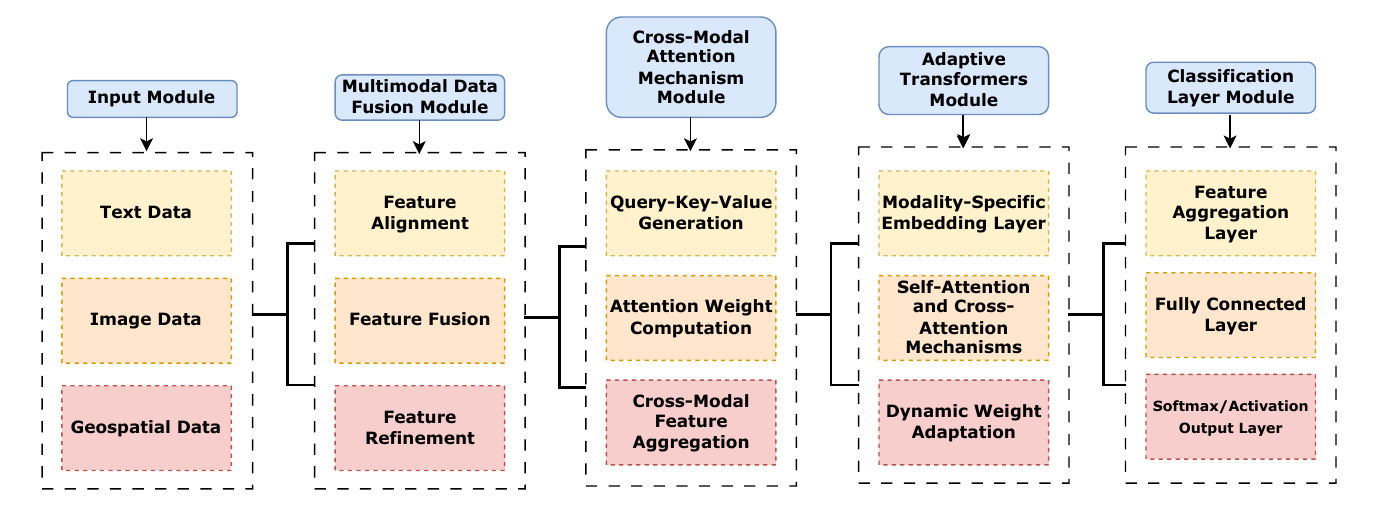}
	\caption{\textbf{DisasterNet-LLM} architecture.}\label{fig:IMRAD}
\end{figure}

Fig. \ref{fig:IMRAD} shows the architecture consists of a hybrid framework combining LLMs (GPT\footnote{{\url{https://huggingface.co/openai-community/openai-gpt}}}) for textual data, CLIP\footnote{{\url{https://huggingface.co/docs/transformers/en/model_doc/clip}}} for image data, and Geospatial Neural Networks (NN-GLS) \cite{zhan2024neural} for spatial data. Below, we present the mathematical formulation of each of the core components of \textbf{DisasterNet-LLM}.

\subsection{Multimodal Data Fusion Module}
Let \( T = \{ t_1, t_2, ..., t_N \} \) denote the set of textual data (social media posts, reports), \( I = \{ i_1, i_2, ..., i_M \} \) the set of image data (disaster-related images), and \( G = \{ g_1, g_2, ..., g_L \} \) the set of geospatial data (locations, coordinates, etc.). The goal of \textbf{DisasterNet-LLM} is to fuse these disparate modalities effectively for disaster prediction and classification. We model the interaction between these modalities. Let \( \mathbf{T} \in \mathbb{R}^{N \times d_t} \) represent the textual embeddings generated by a pretrained LLM such as GPT, where \( d_t \) is the dimensionality of the text embeddings. Let \( \mathbf{I} \in \mathbb{R}^{M \times d_i} \) represent the image embeddings extracted using the CLIP model, where \( d_i \) is the dimensionality of the image embeddings. CLIP is trained to map images and text to a shared representation space, enabling direct comparison and fusion. Let \( \mathbf{G} \in \mathbb{R}^{L \times d_g} \) represent the geospatial embeddings obtained from a NN-GLS, where \( d_g \) is the dimensionality of the geospatial features. A NN-GLS can effectively capture spatial relationships and map them into a feature space compatible with text and image data. The multimodal embedding space is constructed by concatenating these individual modality embeddings as shown in Equation (1):
\[
\mathbf{X} = [\mathbf{T}; \mathbf{I}; \mathbf{G}] \in \mathbb{R}^{(N+M+L) \times d} \tag{1}
\]
where \( \mathbf{X} \) represents the combined multimodal feature space, with \( d \) being the unified dimension.

\subsection{Cross-Modal Attention Mechanism Module}
To model the interactions between the different modalities, we employ a cross-modal attention mechanism based on the Transformer\footnote{{\url{https://huggingface.co/docs/transformers/en/index}}} architecture. The attention mechanism computes pairwise attention scores\footnote{{Pairwise attention scores measure the relevance of one element to another within or across modalities, influencing information weighting and fusion.}} between elements of each modality and across modalities. Let \( Q, K, V \in \mathbb{R}^{(N+M+L) \times d} \) represent the query, key, and value matrices, respectively. The attention mechanism can be described as according to Equation (2):
\[
\text{Attention}(\mathbf{Q}, \mathbf{K}, \mathbf{V}) = \text{softmax}\left( \frac{\mathbf{Q} \mathbf{K}^T}{\sqrt{d}} \right) \mathbf{V} \tag{2}
\]
In the case of cross-modal attention, the queries come from one modality, while the keys and values come from a combination of all modalities. Let \( \mathbf{Q}_t, \mathbf{K}_t, \mathbf{V}_t \in \mathbb{R}^{N \times d} \) be the textual query, key, and value, and similarly define \( \mathbf{Q}_i, \mathbf{K}_i, \mathbf{V}_i \) for images, and \( \mathbf{Q}_g, \mathbf{K}_g, \mathbf{V}_g \) for geospatial data. The output of the cross-modal attention layer can be formulated as according to Equation (3):
\[
\mathbf{A} = \text{Attention}(\mathbf{Q}_t, [\mathbf{K}_t, \mathbf{K}_i, \mathbf{K}_g], [\mathbf{V}_t, \mathbf{V}_i, \mathbf{V}_g]) \tag{3}
\]
This results in an attention matrix \( \mathbf{A} \in \mathbb{R}^{N \times (N+M+L)} \), which captures the pairwise attention scores across text, image, and geospatial data.

\subsection{Adaptive Transformers Module}

The adaptive transformer layer enables dynamic fusion of information from the cross-modal attention scores. In the traditional Transformer model, the attention weights are fixed based on the learned attention mechanism. However, in \textbf{DisasterNet-LLM}, the attention weights are adapted in real-time based on the incoming data. The adaptive weights are computed by applying a gating mechanism as shown in Equation (4):
\[
\mathbf{W}_{adapt} = \sigma\left( \mathbf{W}_a \mathbf{X} + \mathbf{b}_a \right) \tag{4}
\]
where \( \sigma \) is the sigmoid activation function, \( \mathbf{W}_a \in \mathbb{R}^{d \times d} \) and \( \mathbf{b}_a \in \mathbb{R}^{d} \) are learned parameters. The output of the adaptive transformer is then given by according to Equation (5):
\[
\mathbf{Y}_{adapt} = \mathbf{W}_{adapt} \mathbf{X} \tag{5}
\]
The adaptive transformer layer ensures that the network focuses on the most relevant features from each modality depending on the task at hand i.e. classification.

\subsection{Classification Layer Module}

For disaster classification, a fully connected layer maps the multimodal features to a predicted label. The output of the adaptive transformer layer is passed through a softmax function to generate the class probabilities as shown in Equation (6):
\[
\hat{y} = \text{softmax}(\mathbf{W}_c \mathbf{Y}_{adapt} + \mathbf{b}_c) \tag{6}
\]
where \( \mathbf{W}_c \in \mathbb{R}^{C \times d} \) and \( \mathbf{b}_c \in \mathbb{R}^{C} \) are the learned parameters, and \( C \) is the number of classes. The model's loss function is defined as the categorical cross-entropy loss according to Equation (7):
\[
\mathcal{L} = - \sum_{c=1}^{C} y_c \log(\hat{y}_c) \tag{7}
\]
where \( y_c \) is the true label for class \( c \), and \( \hat{y}_c \) is the predicted probability for class \( c \).



\begin{table*}
\scriptsize
    \centering
    \caption{Comparison of \textbf{DisasterNet-LLM} with SOTA on  \cite{niloy2021novel} and \cite{alam2023medic} Datasets.}
    \label{tab:comparison}
    \begin{tabular}{l S[table-format=2.2] S[table-format=2.2] S[table-format=1.2] S[table-format=2.2] S[table-format=2.2] S[table-format=1.2] S[table-format=1.2] S[table-format=1.2]}
        \toprule
        \textbf{Learning Model} & \textbf{Accuracy} & \textbf{F1 Score} & \textbf{AUC} & \textbf{Precision} & \textbf{Recall} & \textbf{BERTScore} & \textbf{MAE} & \textbf{RMSE} \\ 
        \midrule
        Bagging \cite{niloy2021novel} & 61.83 & 59.40 & 0.67 & 60.20 & 57.50 & {N/A} & 0.31 & 0.47 \\ 
        Decision Tree \cite{niloy2021novel} & 44.98 & 42.00 & 0.52 & 43.50 & 40.80 & {N/A} & 0.45 & 0.63 \\ 
        Random Forest \cite{niloy2021novel} & 64.10 & 62.10 & 0.70 & 63.50 & 61.20 & {N/A} & 0.28 & 0.42 \\ 
        K-Nearest Neighbors \cite{niloy2021novel} & 35.67 & 33.50 & 0.48 & 34.80 & 32.10 & {N/A} & 0.50 & 0.72 \\ 
        SVM \cite{niloy2021novel} & 72.52 & 70.40 & 0.78 & 71.80 & 69.00 & {N/A} & 0.21 & 0.34 \\ 
        Linear SVM (w/ SGD) \cite{niloy2021novel} & 66.08 & 63.50 & 0.72 & 65.20 & 62.10 & {N/A} & 0.26 & 0.39 \\ 
        Logistic Regression (w/ SGD) \cite{niloy2021novel} & 65.49 & 63.00 & 0.71 & 64.90 & 61.70 & {N/A} & 0.27 & 0.40 \\ 
        Bagging \cite{alam2023medic} & 43.22 & 40.50 & 0.50 & 42.30 & 38.70 & {N/A} & 0.46 & 0.68 \\ 
        Decision Tree \cite{alam2023medic} & 33.55 & 31.20 & 0.44 & 32.80 & 29.50 & {N/A} & 0.53 & 0.78 \\ 
        Random Forest \cite{alam2023medic} & 46.02 & 43.80 & 0.55 & 44.90 & 42.00 & {N/A} & 0.41 & 0.59 \\ 
        K-Nearest Neighbors \cite{alam2023medic} & 41.86 & 39.70 & 0.50 & 40.80 & 38.50 & {N/A} & 0.48 & 0.70 \\ 
        SVM \cite{alam2023medic} & 54.46 & 51.80 & 0.62 & 53.50 & 50.10 & {N/A} & 0.35 & 0.51 \\ 
        Linear SVM (w/ SGD) \cite{alam2023medic} & 43.66 & 41.20 & 0.50 & 42.80 & 39.50 & {N/A} & 0.45 & 0.67 \\ 
        Logistic Regression (w/ SGD) \cite{alam2023medic} & 43.27 & 41.00 & 0.49 & 42.60 & 39.20 & {N/A} & 0.46 & 0.68 \\ \midrule
        \cellcolor{yellow!25}\textbf{DisasterNet-LLM (Proposed Model)} & \cellcolor{yellow!25}\textbf{89.50} & \cellcolor{yellow!25}\textbf{88.00} & \cellcolor{yellow!25}\textbf{0.92} & \cellcolor{yellow!25}\textbf{89.00} & \cellcolor{yellow!25}\textbf{87.50} & \cellcolor{yellow!25}\textbf{0.88} & \cellcolor{yellow!25}\textbf{0.12} & \cellcolor{yellow!25}\textbf{0.18} \\ 
        \bottomrule
    \end{tabular}
\end{table*}

\section{Experiments}
\subsection{Datasets Analysis}
The \textbf{DisasterNet-LLM} model is evaluated on four datasets. The first dataset, proposed by Niloy et al. \cite{niloy2021novel}, contains 13,720 images of natural disasters, categorized into twelve types such as human damage, infrastructure, urban fire, and sea disasters. The images, sourced from platforms like Google, focus on events such as the Japan tsunami and the Kerala (India) floods. The data is split into 70\% training, 20\% validation, and 10\% testing, with 12 classes. Some categories, like earthquakes, have fewer images, making classification more challenging. The second dataset, MEDIC \cite{alam2023medic}, consists of 71,198 images from sources like CrisisMMD\footnote{{\url{https://crisisnlp.qcri.org/crisismmd}}}, AIDR\footnote{{\url{https://aidr.qcri.org/}}}, and DMD\footnote{\href{https://github.com/husseinmozannar/multimodal-deep-learning-for-disaster-response}{https://github.com/husseinmozannar/\break multimodal-deep-learning-for-disaster-response}}, categorized into seven classes: earthquake, fire, flood, hurricane, landslide, other disaster, and non-disaster. After filtering out mislabeled or irrelevant images, the dataset was reduced to 38,094 images, with a 70\% training, 20\% validation, and 10\% testing split. The third dataset, the Meteorological Raster Dataset (ERA5) \cite{hersbach2020era5}, is sourced from the European Centre for Medium-Range Weather Forecasts and includes four variables: "2m temperature", "10m u-component of wind", "10m v-component of wind", and "total precipitation". The dataset spans July 2015 to June 2023, with a spatial resolution of 27.75 km × 27.75 km. The data is normalized to the range [0, 1]. The fourth dataset, the environmental news \cite{li2024cllmate} dataset , consists of 41,088 news articles related to environmental issues, primarily from East Asian regions. It spans from July 2015 to June 2023, and includes titles, content, statistics, dates, publishers, and media types (web and publications), excluding internet-based media reliant on aggregating news reports. 

\subsection{Hyperparameters}
The hyperparameters used for training include a learning rate of \(10^{-4}\), a batch size of 32, and 50 epochs. The Adam optimizer with weight decay of 0.01 is used, along with a dropout rate of 0.2. The input image resolution is set to 224 × 224. Each dataset is pre-processed carefully, and the model’s performance is evaluated using several metrics, including accuracy\footnote{Proportion of correct predictions over total predictions.}, F1 score\footnote{Harmonic mean of precision and recall.}, AUC\footnote{Area under the ROC curve, measuring classifier's ability.}, precision\footnote{Proportion of correct positive predictions out of total positive predictions.}, recall\footnote{Proportion of correct positive predictions out of actual positives.}, BERTScore\footnote{Uses BERT embeddings for semantic similarity of text.}, Mean Absolute Error (MAE)\footnote{Average of absolute differences between predicted and actual values.}, and Root Mean Squared Error (RMSE)\footnote{Square root of average squared differences between predicted and actual values.}. \textit{\textbf{Note:}} The baseline lacked a number of measures; for instance, BERTScore was present but the others were absent, or vice versa. Or just one metric is displayed. Consequently, we rerun those baseline models in order to compute those missing measures.

\subsection{Comparison with State-of-the-Art}
\begin{table*}
\scriptsize
    \centering
    \caption{Comparison of \textbf{DisasterNet-LLM} with SOTA models on \cite{hersbach2020era5} and \cite{li2024cllmate} datasets.}
    \label{tab:comparison_sota}
    \begin{tabular}{l S[table-format=2.2] S[table-format=2.2] S[table-format=1.2] S[table-format=2.2] S[table-format=2.2] S[table-format=1.2] S[table-format=1.2] S[table-format=1.2]}
        \toprule
        \textbf{Learning Model} & \textbf{Accuracy} & \textbf{F1 Score} & \textbf{AUC} & \textbf{Precision} & \textbf{Recall} & \textbf{BERTScore} & \textbf{MAE} & \textbf{RMSE} \\ 
        \midrule
        Top-1 Sim \cite{li2024cllmate} & 71.37 & 69.50 & 0.75 & 70.80 & 68.10 & 0.70 & 0.22 & 0.36 \\ 
        LLaVA Tuning \cite{li2024cllmate} & 69.84 & 68.00 & 0.73 & 69.30 & 67.10 & 0.68 & 0.25 & 0.39 \\ 
        CLLMate \cite{li2024cllmate} & 73.56 & 72.00 & 0.78 & 73.10 & 71.30 & 0.73 & 0.20 & 0.32 \\ \midrule
        \cellcolor{yellow!25}\textbf{DisasterNet-LLM (Proposed Model)} & \cellcolor{yellow!25}\textbf{89.50} & \cellcolor{yellow!25}\textbf{88.00} & \cellcolor{yellow!25}\textbf{0.92} & \cellcolor{yellow!25}\textbf{89.00} & \cellcolor{yellow!25}\textbf{87.50} & \cellcolor{yellow!25}\textbf{0.88} & \cellcolor{yellow!25}\textbf{0.12} & \cellcolor{yellow!25}\textbf{0.18} \\ 
        \bottomrule
    \end{tabular}
\end{table*}

Table~\ref{tab:comparison} provides a comparative analysis of the proposed \textbf{DisasterNet-LLM} model against existing State-of-the-Art (SOTA) methods from \cite{niloy2021novel} and \cite{alam2023medic} on two disaster datasets. The results demonstrate the superior performance of \textbf{DisasterNet-LLM} across all evaluation metrics. The proposed model achieves an accuracy of 89.50\% and an F1 score of 88.00\%, significantly outperforming the best-performing existing method, which is the SVM model from \cite{niloy2021novel} with an accuracy of 72.52\% and an F1 score of 70.40\%. The AUC value of 0.92 indicates \textbf{DisasterNet-LLM} exceptional ability to distinguish between classes, while its precision and recall values, both exceeding 87\%, highlight its effectiveness in minimizing false positives and false negatives. Additionally, the model's achieves a BERTScore of 0.88, demonstrating its strength in capturing semantic similarities in text data. Furthermore, the lower MAE and RMSE values of 0.12 and 0.18, respectively, reflect the model's robustness and consistency in predictions.

The Table \ref{tab:comparison_sota} compares the performance of the proposed \textbf{DisasterNet-LLM} with SOTA models from \cite{li2024cllmate} on other two datasets.  \textbf{DisasterNet-LLM} achieves the highest accuracy of 89.50\%, compared to 71.37\% by Top-1 Sim, 69.84\% by LLaVA Tuning, and 73.56\% by CLLMate. Similarly, the F1 score, AUC, and BERTScore of the proposed model are superior, showcasing its robust classification and prediction capabilities. Additionally, \textbf{DisasterNet-LLM} achieves significant reductions in error metrics, with the lowest MAE of 0.12 and RMSE of 0.18, highlighting its precision and reliability.

\subsection{Additional Experiments}
\begin{figure}
	\centering
	\includegraphics[width=0.45\textwidth, height=0.2\textwidth]{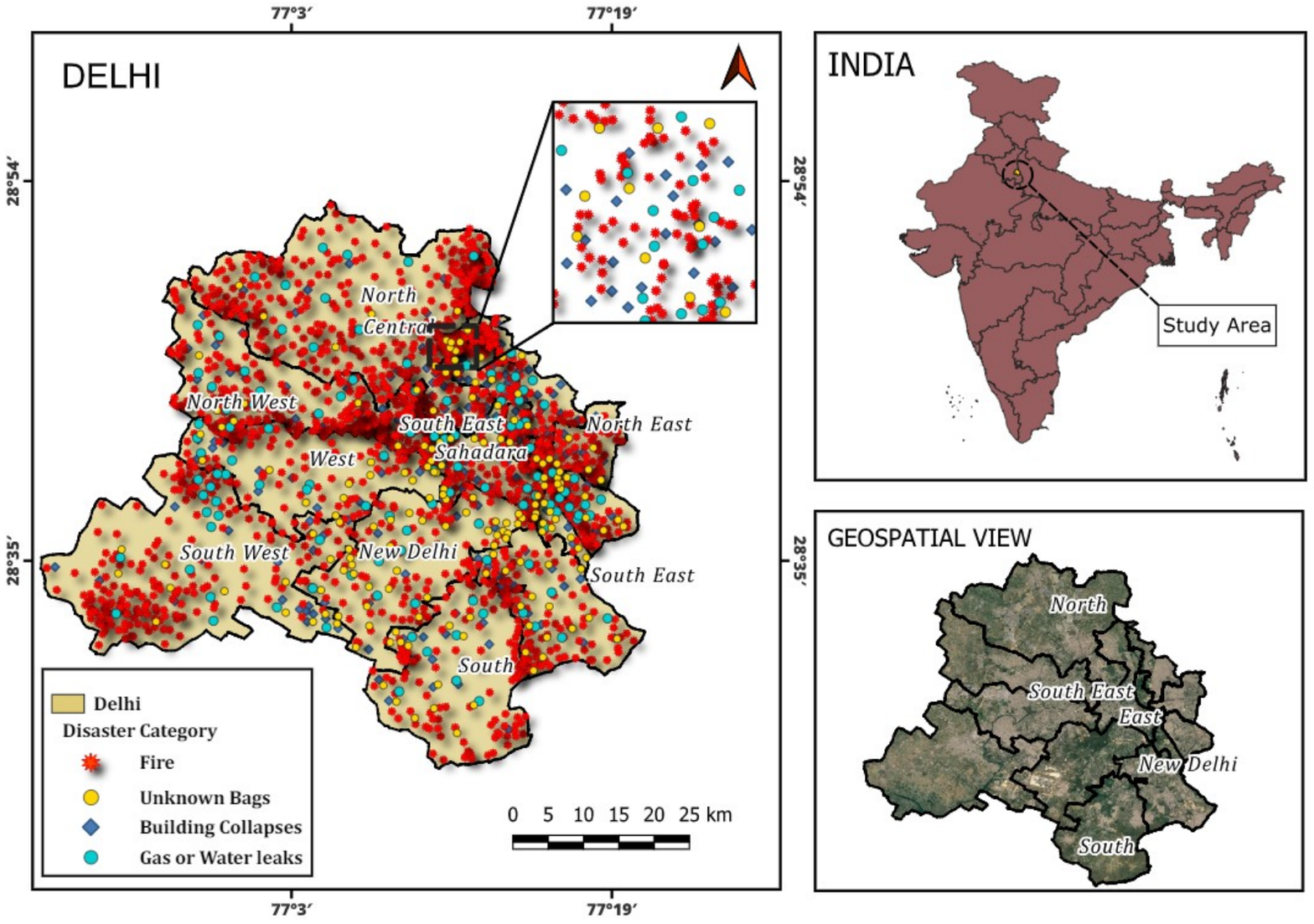}
	\caption{Base map of Delhi}\label{fig:delhi}
\end{figure}

To evaluate \textbf{DisasterNet-LLM} performance on real-world urban risk scenarios, we conducted experiments using a dataset collected by the National Institute of Disaster Management-Delhi (NIDM)\footnote{{\url{https://nidm.gov.in/}}}, spanning March 2022 to May 2024 of Delhi (India) as shown in Fig. \ref{fig:delhi}, which includes records of urban disasters such as fire, unknown bags, building collapses, and gas or water leaks. The dataset comprises temporal and spatial variables, enabling analysis of urban safety trends. The 2022 data revealed fire disasters and gas leaks as dominant, while the 2023 data showed rising cases of mystery bags and vehicle-related events, likely due to improved reporting. Seasonal spikes in building collapses highlighted monsoon-related vulnerabilities. Early 2024 data indicated increased gas and water leaks, suggesting rising utility demands or better detection systems. Preprocessing addressed missing values and underreporting. \textbf{DisasterNet-LLM} demonstrated high performance as shown in Table \ref{tab:urban_risk_results}, achieving 91.65\% accuracy and 90.95\% F1 score overall. Gas and water leaks had the highest metrics, while fire disasters also scored well, showcasing the model’s reliability and utility for urban disaster management.
\begin{table}[hbt]
\scriptsize
    \centering
    \caption{Performance of \textbf{DisasterNet-LLM} on Urban Risk Dataset (March 2022–May 2024).}
    \label{tab:urban_risk_results}
    \begin{tabular}{p{2cm} p{0.8cm} p{1cm} p{0.8cm} p{0.7cm} p{1cm}}
        \toprule
        \textbf{Disaster Category} & \textbf{Accuracy} & \textbf{F1 Score} & \textbf{Precision} & \textbf{Recall} & \textbf{BERTScore} \\ 
        \midrule
        Fire & 92.34 & 91.80 & 91.90 & 91.70 & 0.89 \\ 
        Unknown Bags & 88.12 & 87.50 & 87.80 & 87.20 & 0.85 \\ 
        Building Collapses & 90.67 & 89.90 & 90.10 & 89.80 & 0.87 \\ 
        Gas or Water Leaks & 93.45 & 92.80 & 92.90 & 92.70 & 0.90 \\ \midrule
        \cellcolor{yellow!25}\textbf{Overall} & \cellcolor{yellow!25}\textbf{91.65} & \cellcolor{yellow!25}\textbf{90.95} & \cellcolor{yellow!25}\textbf{90.92} & \cellcolor{yellow!25}\textbf{90.85} & \cellcolor{yellow!25}\textbf{0.88} \\ 
        \bottomrule
    \end{tabular}
\end{table}
\vspace{-1em}

\section{Conclusion and Future Works}

In this study, we proposed \textbf{DisasterNet-LLM}, a robust multimodal framework for disaster prediction. The model demonstrated superior performance across multiple metrics, outperforming existing SOTA approaches. Its ability to integrate textual and visual modalities effectively enables comprehensive decision-making in disaster management. For future work, we aim to enhance the model’s scalability for diverse geographical regions, and integrating domain-specific knowledge from meteorological experts could further improve accuracy and applicability. Additionally, due to page constraints, we missed including the category-wise classification of disasters in the dataset; this will be addressed in future work.

\small
\bibliographystyle{IEEEtranN}
\bibliography{references}

\end{document}